\documentclass[runningheads]{llncs}
\usepackage[T1]{fontenc}
\usepackage{tabularx}
\newcolumntype{Y}{>{\centering\arraybackslash}X}
\usepackage{amssymb}
\usepackage{amsmath}
\usepackage{comment}
\usepackage{float}
\usepackage[misc]{ifsym} 
\usepackage{subfigure}
\usepackage{cite}
\usepackage[colorlinks,bookmarksopen,bookmarksnumbered,citecolor=blue, linkcolor=blue, urlcolor=blue]{hyperref}
\usepackage{graphicx}
\begin{document}
\title{LSM-YOLO: A Compact and Effective ROI Detector for Medical Detection}
%
%
\author{Zhongwen Yu\inst{1} \and
Qiu Guan\inst{1} \textsuperscript{(\Letter)} \and
Jianmin Yang\inst{2} \and
Zhiqiang Yang\inst{1} \and
Qianwei Zhou\inst{1} \and
Yang Chen\inst{3} \and
Feng Chen\inst{4}}
\authorrunning{Z. Yu et al.}
%
\institute{College of Computer Science and Technology, Zhejiang University of Technology, Hangzhou, China\\
\email{vincentyu67373@gmail.com} \\
\email{gq@zjut.edu.cn} \and
Zhejiang College of Sports, Hangzhou, China \and
SouthEast University, School of Computer Science and Engineering,\\
Nanjing, China \and
The First Affiliated Hospital, Zhejiang University School of Medicine,\\
Hangzhou, China}
\maketitle              
\begin{abstract}
In existing medical Region of Interest (ROI) detection, there lacks an algorithm that can simultaneously satisfy both real-time performance and accuracy, not meeting the growing demand for automatic detection in medicine. Although the basic YOLO framework ensures real-time detection due to its fast speed, it still faces challenges in maintaining precision concurrently. To alleviate the above problems, we propose a novel model named Lightweight Shunt Matching-YOLO (LSM-YOLO), with Lightweight Adaptive Extraction (LAE) and Multipath Shunt Feature Matching (MSFM). Firstly, by using LAE to refine feature extraction, the model can obtain more contextual information and high-resolution details from multiscale feature maps, thereby extracting detailed features of ROI in medical images while reducing the influence of noise. Secondly, MSFM is utilized to further refine the fusion of high-level semantic features and low-level visual features, enabling better fusion between ROI features and neighboring features, thereby improving the detection rate for better diagnostic assistance. Experimental results demonstrate that LSM-YOLO achieves 48.6\% AP on a private dataset of pancreatic tumors, 65.1\% AP on the BCCD blood cell detection public dataset, and 73.0\% AP on the Br35h brain tumor detection public dataset. Our model achieves state-of-the-art performance with minimal parameter cost on the above three datasets. The source codes are at: \url{https://github.com/VincentYuuuuuu/LSM-YOLO}.

\keywords{Medical image detection \and YOLO \and Lightweight \and Adaptive Feature Extraction \and Multipath Feature Fusion.}
\end{abstract}
\section{Introduction}
Medical imaging techniques such as Computed Tomography (CT) and Magnetic Resonance Imaging (MRI) are widely used in radiological examinations \cite{1,2}. The corresponding diagnoses, such as tumor screening and blood imaging analysis, which predominantly rely on senior doctors visually analyzing the images. There is a lack of tools and methods for automatic detection of Region of Interest (ROI) in medical images. Existing analyses based on medical imaging mostly focus on medical image segmentation, with few mature detection outcomes being applied in clinical practice. The sizes of lesions in medical imaging are not uniform, and lesions are difficult to detect in the early stages. Additionally, there is a scarcity of medical ROI detection models. What's more, The small scale of medical datasets is not conducive to data-driven methods like deep learning. Therefore, our work hopes to expand in this field, combining automatic detection with medicine, aiming to provide a lightweight, easily deployable, efficient medical object detection model.\\
\indent Nowadays, mainstream detectors fall into two categories: CNN-based, exemplified by YOLO \cite{v1,v5,v6,v7,v8,gold,v9,v10}, and Transformer-based, illustrated by DETR \cite{detr,dino,rt,defo}. The YOLO framework stands out for its remarkable balance between speed and accuracy, enabling fast and reliable object detection in images. Despite DETR's excellent accuracy on large datasets like MS-COCO, its high computational cost and parameter count hamper its comprehensive performance, especially in specific sectors with smaller datasets, like medicine, where YOLO excels. It's this superiority that motivates YOLO's use in this work.\\
\indent In medical imaging, the ROI for detection are relatively smaller, as lesions often only occur in certain parts of an organ in CT or MRI scans. And the relationship between the ROI and surrounding areas is crucial. How to properly handle the relationship between the two is an important issue. Typically, functions deployed on medical devices should be lightweight and should not consume too much resource, lowering operational costs while improving the quality of medical services. This is why we aim to achieve lightweight models.\\
\indent To address the aforementioned challenges, we have developed a lightweight and easily deployable LSM-YOLO framework to contribute to medical image detection. Initially, we introduced the Receptive-Field Attention Convolutional operation (RFAConv) \cite{rfa}, utilizing a spatial attention mechanism to share convolutional kernel parameters. This means in the initial stages of feature extraction, the model learns the similarities and differences between the object area and its surroundings. To extract multi-scale feature maps, we designed the Lightweight Adaptive Extraction (LAE) to allow downsampling on the basis of retaining as much information-rich features as possible, which not only circumvents the issue of high computational demand associated with traditional convolutions but also dynamically extracts features during multi-scale sampling, allowing the model's attention to focus more significantly on the object areas. After obtaining multi-scale feature maps, naturally, we provide a rational feature fusion mechanism—Multipath Shunt Feature Matching (MSFM). This allows the model to focus more on coordinates with greater weight values during the feature fusion process, thereby learning key information such as tumor characteristics and locations. Through the methods described above, we can obtain richer contextual and high-resolution information, which enhances the detection accuracy and generalizability of the model. Furthermore, to address the issue of numerous small objects in medical imaging, we optimize the output head, specifically extending the Path Aggregation-Feature Pyramid Network (PA-FPN) \cite{pafpn} structure to improve detection capabilities for small objects, using four detection heads to complete the final output task.

\section{Related Work}
\subsection{Medical ROI Detection}
Medical ROI detection plays a crucial role in improving diagnosis accuracy, treatment planning, and surgical interventions. By automatically identifying and localizing targets of ROI, it helps radiologists and doctors in detecting early signs of diseases, such as cancer, enabling prompt intervention and better patient treatments. \\
\indent Aiming at different organ regions, there are many works utilizing deep learning methods to achieve automatic medical ROI detection.Kang et al. \cite{rcs} proposed RCS-YOLO to address the detection of brain tumors. Ahmed et al. \cite{ab} introduced the YOLO model into wrist abnormality detection. DeGPR \cite{degpr} is a model focused on cell detection and counting, which can assist other object detectors. Huynh et al.\cite{ac} utilized object detection models for the automatic detection of acne skin diseases. CircleNet \cite{cir} is a model designed for detecting ball-shaped biomedical objects, such as glomeruli and cell nuclei, the author also proposes the use of circular bounding boxes for detecting spherical objects in medical tasks, aiming to better accommodate detection tasks with unique shapes. In the field of medical image processing, Shamshad et al. \cite{tr} conducted a comprehensive review of the various applications of Transformers, including classification, segmentation, detection, reconstruction, and registration. This systematic compilation is highly commendable.

\subsection{Multi-scale Features for Object Detection}
Introducing multi-scale features into object detection models is essential. Typically, larger features capture the textural details of smaller objects, while smaller features contain the semantic attributes of larger objects. Relying solely on single-scale features will omit detail information. For example, in an abdominal CT image, the size difference between the stomach and pancreas is significant. With only single-scale feature assistance, it is difficult for detectors to fully learn the unique features of the stomach and pancreas, possibly leading the model to recognize only the stomach without knowing what the pancreas looks like. \\
\indent Balancing high-level semantic information with low-level visual information has always been a focus of numerous works. In this context, the Feature Pyramid Network (FPN) \cite{fpn} structure has been widely adopted due to its method of cross-scale connection and fusion of multi-scale features. Subsequent work has optimized the FPN to further address the above issue. Path Aggregation Network (PAN) \cite{pafpn} enhances the integration of features at various levels through a bottom-up pathway. Bidirectional Feature Pyramid Network (BiFPN) \cite{bifpn}, a weighted bidirectional feature pyramid network, introduces learnable weights to apply multi-scale feature fusion from top to bottom and bottom to top repetitively. Unlike inter-layer interactions, Centralized Feature Pyramid (CFP) \cite{cfp} focuses on intra-layer features to capture global long-distance dependencies. Additionally, Asymptotic Feature Pyramid Network (AFPN) \cite{afpn} progresses to high-level features by fusing the low-level information of two adjacent layers, supporting interactions between non-adjacent layers. However, the introduction of multi-scale features into the model inevitably increases the computational cost. Deformable DETR \cite{defo} achieves a good fusion of multi-scale features at the expense of computational resources. Nevertheless, excessive information interchange might lead to the loss of low-level information and poses a challenge in balancing efficiency, accuracy, and lightweight design.

\section{Methods}
\begin{figure}
\centering
\includegraphics[width=0.9\textwidth]{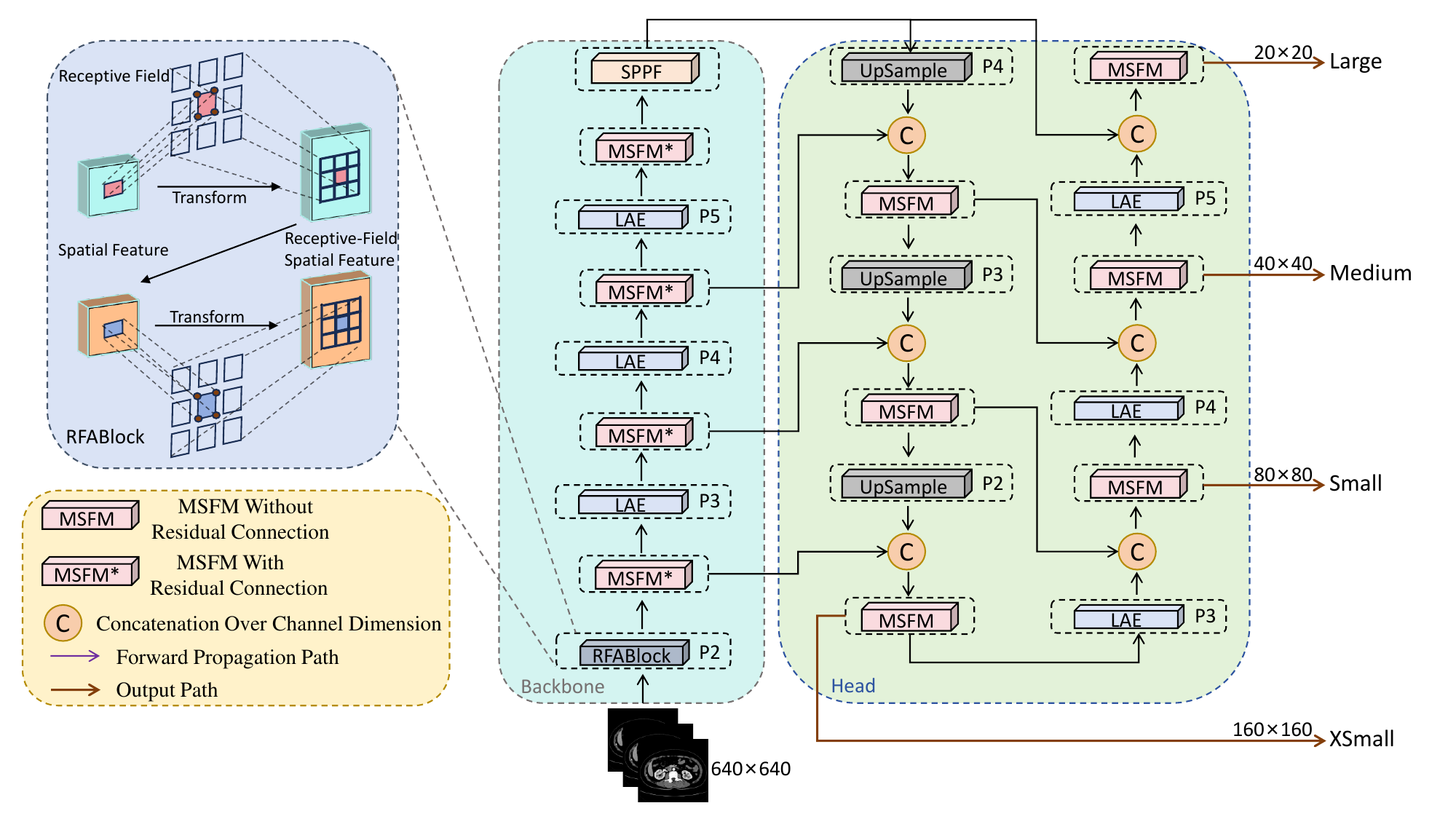}
\caption{Overview of the network architecture of LSM-YOLO. 1) LAE as a module to extract multi-scale feature maps; 2) MSFM as a module to refine and fuse high-level semantic and low-level spatial features.} \label{Fig.1}
\end{figure}

The architecture of the proposed LSM-YOLO network is shown in Fig.\ref{Fig.1}. The network mainly extracts features in the backbone, integrates features in the head section, and combines with the four output heads to output.

\subsection{Lightweight Adaptive Extraction}
\begin{figure}
\centering
\includegraphics[width=0.9\textwidth]{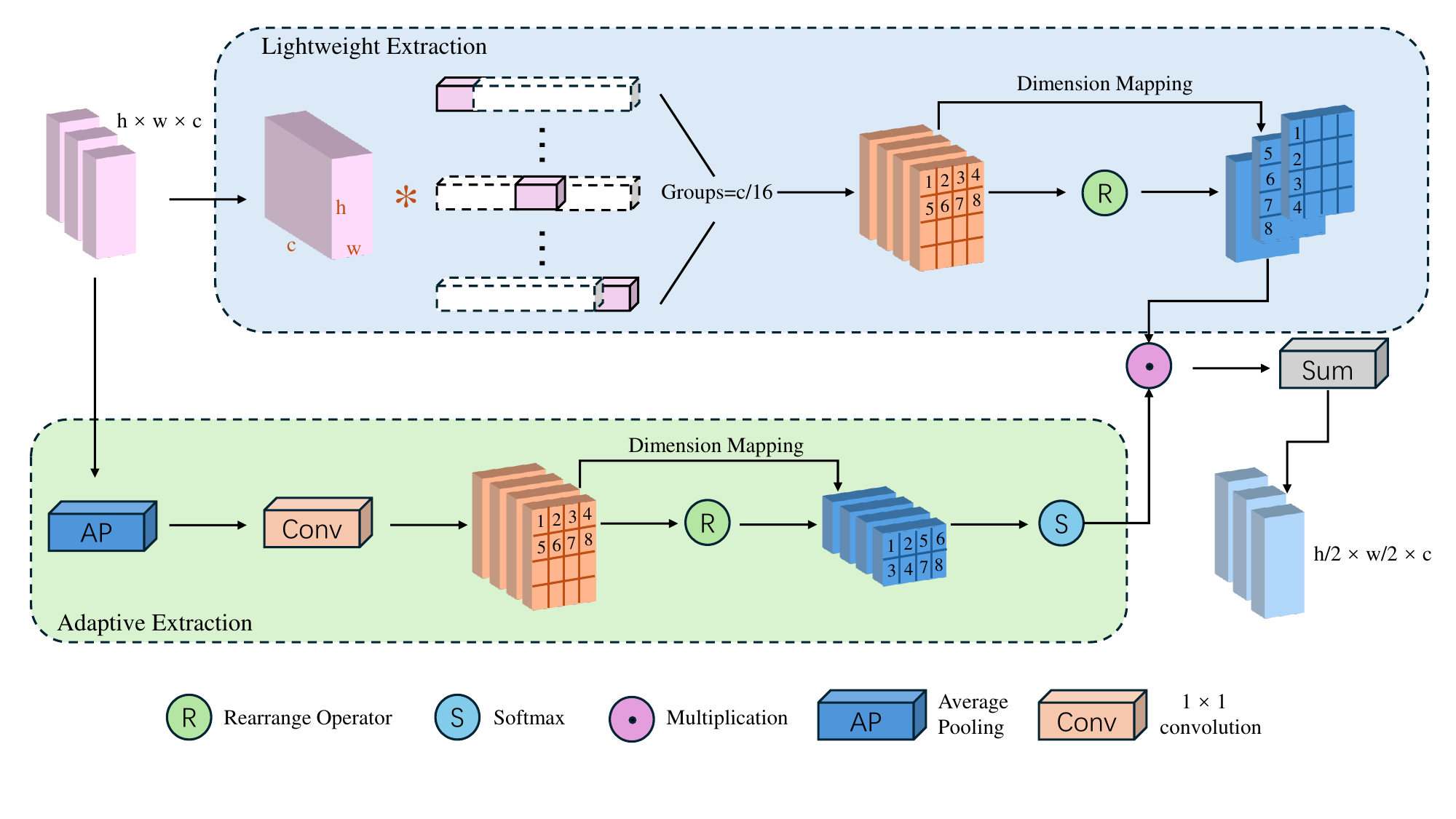}
\caption{The structure of Lightweight Adaptive Extraction (LAE) module.} \label{Fig.2}
\end{figure}

In multi-scale feature extraction, compared to the traditional convolution methods, LAE significantly reduces the number of parameters and computational cost, while also extracting features with richer semantic information. Fig.\ref{Fig.2} shows the structural schematic diagram of LAE.
\\ \indent In the process of traditional convolution, pixel information at the edges and corners is lost, but these local information is particularly important in medical imaging, as it represents the implicit information between the ROI targets and the surrounding areas. Additionally, there is a difference in information between adjacent pixels of the feature map, pixels containing the object have higher information entropy than their neighboring pixels, so we hope to preserve the pixels with higher information content during the sampling process.\\
\indent Considering the local nature of convolution operations, difficult to capture global information while also involving complex computations. we referred to the concept of Focus \cite{v5}, which uses slice operation to separate sampling by rows and columns. This achieves the aim of sampling while reducing information loss, essentially concentrating the feature map's height and width information onto the channel level. However, we did not directly adopt the slice operation due to its high computational cost, which contradicts our goal of lightweight processing. \\
\indent We designed two parallel branches with shared parameters, adopting the concept of group convolution to map the input to the output dimensions efficiently with low parameters. By utilizing group convolution with N groups, the number of parameters is reduced to 1/N compared to traditional convolution.\\
\indent Each LAE unit achieves a four-fold downsampling, which means scaling both height and width by a factor of two. In order to alleviate the loss of edge information during sampling process, after saving feature map's height and width information into channels, the feature map's dimensions change from four-dimensional (batch, channel, height, weight) to five-dimensional (batch, channel, height, weight, n), where 'n' represents the sampling factor. Adaptive Extraction path exchanges information via average pooling and convolution. Essentially, downsampling on this path recombines the feature map according to four adjacent pixels (such as the top left corner four pixels), with their respective weights expressed through softmax, also transforming the dimensions into five-dimensional. On 'n' dimension, adaptive weights are combined with the other branch. This method can be understood as implicitly including global information at the channel level during the transition from high-resolution to low-resolution information. Overall, the two branches are responsible for concentrating height and width information into the channels and calculating the corresponding information weights, respectively. Our LAE module is parameter-free when used, facilitating the plug-and-play transfer.

\subsection{Multipath Shunt Feature Matching}
\begin{figure}
\centering
\includegraphics[width=0.9\textwidth]{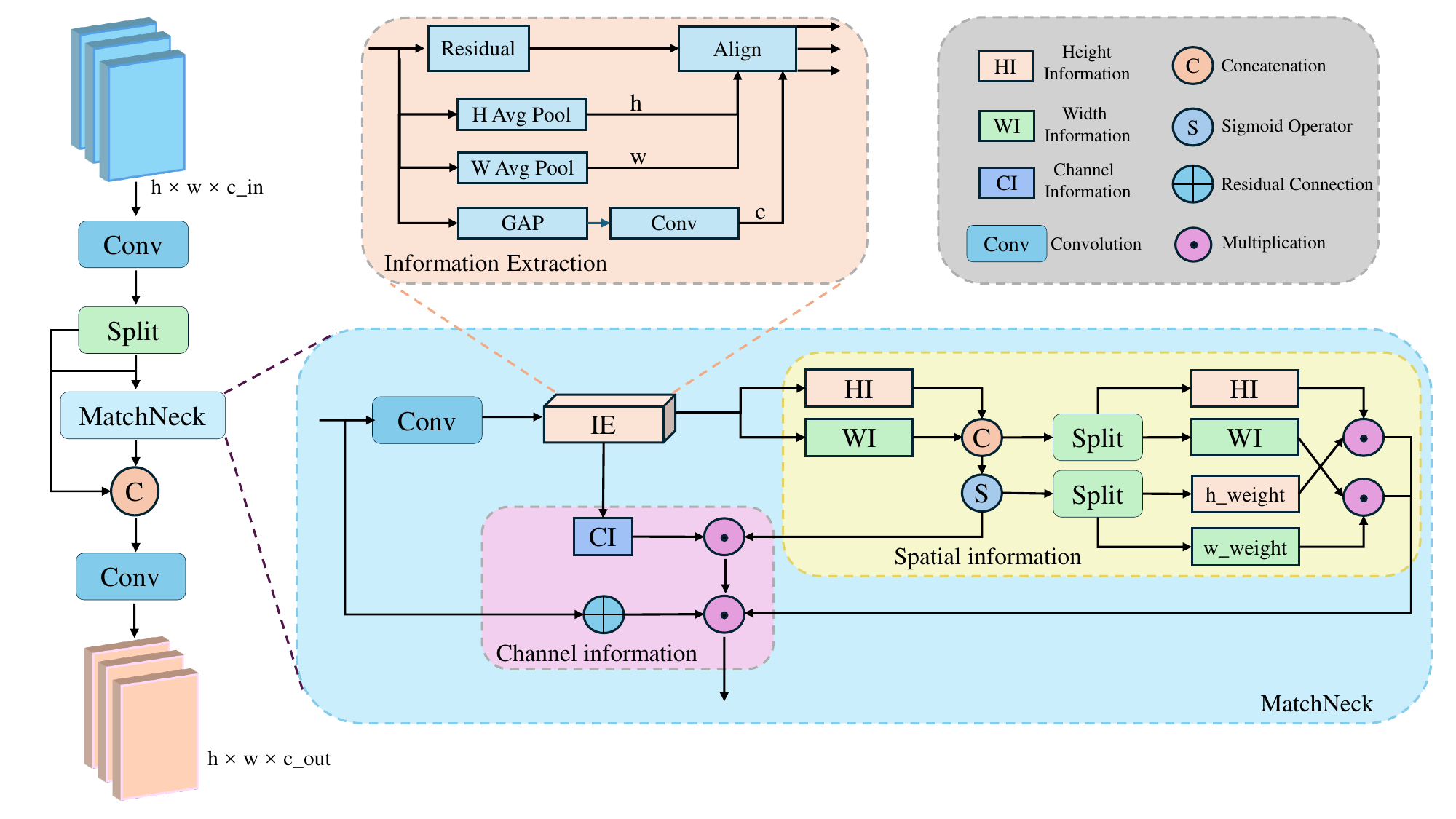}
\caption{The structure of Multipath Shunt Feature Matching (MSFM) module.} \label{Fig.3}
\end{figure}

In multi-scale feature fusion, we aim to break away from the traditional approach of relying solely on the interchange of channel information. This is because the multi-scale features extracted represent the mapping of objects of various sizes on the feature maps. Especially in medical imaging, such as tumor detection, the presence of a tumor is often associated with invasive phenomena and is characterized by multiple occurrences and a high tendency to metastasize. This is reflected in the feature map by a high degree of correlation in both the spatial and channel dimensions. The exchange of channel information can enhance the model’s ability to capture visual information. However, a significant amount of semantic information is concentrated spatially, and the interaction of spatial information can assist in the detection of ROI with different scales. For instance, feature maps with low-resolution yet high semantic information are inherently used to predict larger targets. If we merge the high-level spatial information with the low-level visual information, it can have a complementary effect, using the global information from a larger receptive field to help the low-level feature maps predict smaller objects. Hence, our proposed MSFM module conducts a comprehensive analysis of spatial and channel information across features from low to high levels. Fig.\ref{Fig.3} shows the structural schematic diagram of MSFM.
\\ \indent The MSFM module follows the shunt concept, where the MatchNeck block is used to enhance the model's ability to represent ROI area features while controlling the number of parameters and computational complexity. The process starts by employing a split operator to divide the information flow, retaining the original features for subsequent residual connections. The MSFM module extracts information across the dimensions of height, width, and channels from the input feature tensor in a computationally efficient manner. Let average pooling and global average pooling operation be denoted as $P_{avg}$ and $P_{gavg}$, the process of extracting $F_{out}^{'}$ from the input feature $F_{in}$ is as follows:
\begin{equation}
	F_{h}=P_{avg}(F_{in}^h), F_{w}=P_{avg}(F_{in}^w), F_{c}=P_{gavg}(F_{in}).
\end{equation}
\begin{equation}
    F_{out}^{'}=Res(Align(F_{h},F_{w},F_{c})).
\end{equation}
Incorporating height and width information into channels after normalization can help cascade global information, thereby capturing common features of ROI sites and their neighborhoods to promote the interaction of contextual information. Meanwhile, the spatial information stream integrated into the channel is preserved as auxiliary weights, which are then passed back to the spatial tensor through a multiplication operator. Let the height and width information after post-processing be $\hat{F_{h}}$ and $\hat{F_{w}}$, with corresponding weights $weight_{h}$ and $weight_{w}$, the processing process of spatial information is as follows: 
\begin{equation}
	\hat{F_{h}},\hat{F_{w}}=Split(Concat(F_{h},F_{w})).
\end{equation}
\begin{equation}
    weight_{h},weight_{w}=Split(Sigmoid(Concat(F_{h},F_{w}))).
\end{equation}
Let the channel information after post-processing be $\hat{F_{c}}$, with the output as $F_{out}$, the process of the channel information branch is as follows:
\begin{equation}
	\hat{F_{c}}=F_{c} \times Sigmoid(Concat(F_{h},F_{w})).
\end{equation}
\begin{equation}
	F_{out}=\hat{F_{c}} \times Res(F_{in}) \times \hat{F_{h}}*weight_{h} \times \hat{F_{w}}*weight_{w}. \\
\end{equation}
After concatenating with the original source information and passing through a 1x1 convolution, the output is then produced. \\
\indent MSFM module has two versions: one with residual connections and one without. In the backbone, we use the version with residual connections, as the primary task at this stage is to extract the main features of the object, providing meaningful feature representations for subsequent stages. The use of residual connections helps to mitigate the problem of vanishing gradients, thus accelerating the convergence of the model. In the head, we employ the version without residual connections. At this stage, the model has already extracted ample feature information, and the task is to analyze these features to predict objects. Therefore, residual connections become redundant here. 

\subsection{Loss Function}
The loss function of LSM-YOLO is divided into a classification branch and a regression branch. The classification branch employs Binary Cross-Entropy Loss (BCE), while the regression branch is divided into Distribution Focal Loss (DFL) \cite{dfl} and SCYLLA-IoU (SIoU) Loss \cite{siou}. 
The overall loss in this paper is a weighted combination of the above three, with each component being proportionally weighted. With $\gamma=0.5$, $\zeta=1.5$, $\eta=7.5$, its definition is as follows:
\begin{equation}
	Loss=\gamma BCE Loss+\zeta DFL+\eta SIoU Loss.
\end{equation}

\section{Experiments and Results}
\subsection{Datasets Details}
To evaluate the proposed LSM-YOLO model, we use three different medical datasets. Our private CT pancreatic tumor dataset consists of serous cystic neoplasm and mucinous cystic neoplasm, with 1173 images in the training set and 309 images in the validation set. We use the MRI brain tumor dataset (Br35H) \cite{br35h}, which consists of a total of 701 images. Out of these, 500 images were used for training, and 201 images were used for validation. Additionally, we also utilize the Blood Cell Count and Detection (BCCD) \cite{bccd} dataset, which includes 292 images in the training set and 72 images in the validation set. 

\subsection{Implementation Details}
For model training and inference, we utilize two NVIDIA GeForce RTX 4090 GPUs. For YOLO-series models, we uniformly set the epoch to 300 and the input image size to 640 × 640. For non-YOLO models, we conduct experiments with the epoch settings specified in their respective original papers. However, for RT-DETR, we set the epoch to 300 because we found that using the original paper's \cite{rt} epoch of 72 did not lead to complete convergence on these datasets. 

\subsection{Results}
We divide the comparison models into YOLO series and end-to-end DETR series, which are currently the two mainstream categories in object detection.\\
\indent Tab.\ref{Tab:1} first compares proposed LSM-YOLO with YOLO-series detectors that are closest in size on the pancreatic tumor dataset. LSM-YOLO achieves fewer parameters and significantly higher accuracy. Further comparing it with larger models, LSM-YOLO still has a clear advantage on the pancreatic tumor dataset. In the comparison with DETR series models, the large number of parameters and computational load of the compared models did not result in positive gains. The reason is that the DETR series models lack the processing of multi-scale features, causing the deeper networks can not obtain sufficient ROI tumor features for predictions.\\
\indent Our proposed LSM-YOLO achieves 48.6\% AP$_{50:95}$ and 60.8\% AP$_{50}$ on the pancreatic tumor dataset, outperforming the state-of-the-art detector among numerous detectors with lower parameter count and computational cost. LSM-YOLO has shown absolute superiority in small, medium and large objects detection.\\
\indent We also experimented with larger-sized models, but the results indicated inferior performance compared to the smaller-sized models. From Tab.\ref{Tab:1}, as the model size increases, the detection accuracy decreases. Therefore, it can be concluded that for smaller-scale medical datasets, smaller-sized models perform better.\\
\begin{table}[t]
	\centering
	\caption{Comparison with the state-of-the-art detectors on pancreatic tumor dataset.}\label{Tab:1}
	\begin{tabular}{l|c|c|c|c|c|c|c}
		\hline\hline\noalign{\smallskip}	
		Model & Params (M) & GFLOPs & AP$_{50:95}$ & AP$_{50}$ & AP$_{S}$ & AP$_{M}$ & AP$_{L}$ \\
		\noalign{\smallskip}\hline\noalign{\smallskip}
        YOLOv6-3.0-N & 4.63 & 11.3 & 35.7 & 49.1 & 11.5 & 41.5 & 42.7 \\
        YOLOv7-T & 6.01 & 13.0 & 34.7 & 48.4 & 12.5 & 45.1 & 35.1 \\
        YOLOv8-N & 3.01 & 8.1 & 39.5 & 53.6 & 13.7 & 49.9 & 40.7 \\
        Gold-YOLO-N & 5.60 & 12.1 & 32.7 & 44.7 & 9.3 & 38.1 & 45.7 \\
        YOLOv9-T & 2.62 & 10.7 & 39.6 & 53.4 & 12.1 & 48.8 & 39.8 \\
        YOLOv10-N & 2.70 & 8.2 & 38.3 & 51.7 & 12.0 & 47.3 & 38.0 \\
        \noalign{\smallskip}\hline\noalign{\smallskip}
		YOLOv6-3.0-S & 18.50 & 45.2 & 30.7 & 40.7 & 8.1 & 34.6 & 44.6 \\
		YOLOv7 & 36.49 & 103.2 & 32.5 & 44.0 & 18.1 & 40.0 & 32.1 \\
		YOLOv8-S & 11.13 & 28.4 & 39.2 & 52.2 & 12.8 & 49.8 & 37.9 \\
		RCS-YOLO & 45.70 & 94.5 & 35.9 & 47.6 & 10.1 & 45.7 & 35.7 \\
		Gold-YOLO-S & 21.51 & 46.0 & 37.0 & 50.6 & 19.7 & 40.7 & 46.3 \\
		YOLOv9-S & 9.60 & 38.7 & 39.9 & 52.8 & 13.0 & 50.9 & 36.9 \\
        YOLOv10-S & 8.04 & 24.4 & 37.3 & 51.8 & 16.5 & 48.3 & 34.8 \\
		\noalign{\smallskip}\hline\noalign{\smallskip}
		DINO-4scale-R50 & 46.60 & 179.4 & 27.2 & 36.8 & 7.7 &50.6 &42.4 \\
		DINO-5scale-R50 & 46.97 & 522.8 & 26.7 & 36.8 & 13.1 &47.5 &44.8 \\
		RT-DETR-R50 & 41.96 & 129.5 & 35.0 & 46.4 & 9.1 & 44.2 & 42.4 \\
		RT-DETR-HGNetv2-L & 31.99 &103.4 &32.2 &42.7 &7.4 & 40.8 & 34.5 \\
		\noalign{\smallskip}\hline\noalign{\smallskip}
		LSM-YOLO (ours) & 2.87 & 12.4 & \bfseries{48.6} & \bfseries{60.8} & \bfseries{21.9} & \bfseries{51.1} & \bfseries{54.5} \\
		\noalign{\smallskip}\hline\hline
	\end{tabular}
\end{table}
\indent Tab.\ref{Tab:2} compares our LSM-YOLO with other detectors on the BCCD blood cell dataset. Our LSM-YOLO achieves 65.1\% AP$_{50:95}$ and 92.7\% AP$_{50}$, demonstrating state-of-the-art performance with relatively low parameter count and computational cost.
It can be seen that on blood cell dataset, which is the dense multi-object detection task, the YOLO series models and the DETR series models perform similarly. The reason is that this type of dataset has a distribution where large targets are easy to detect but multiple overlapping targets are difficult to detect. Both types of models do not perform well in detecting the latter. However, our LSM-YOLO, due to its effective feature fusion mechanism, is sensitive to the regions surrounding the targets and better captures potential ROI area features, thus achieving a better detection rate. Compared with RT-DETR-R50, LSM-YOLO reduces the number of parameters by 93.2\% and the computational amount by 90.4\%, while significantly improves accuracy. Experimental results highlight that LSM-YOLO has a clear advantage in the overall AP metrics. While ensuring a high detection rate for easy-to-detect large targets, it also demonstrates excellent detection performance for small and medium targets with overlapping and coverage.\\
\begin{table}[t]
	\centering
	\caption{Comparison with the state-of-the-art detectors on BCCD dataset.}\label{Tab:2}
	\begin{tabular}{l|c|c|c|c|c|c|c}
		\hline\hline\noalign{\smallskip}	
		Model & Params (M) & GFLOPs & AP$_{50:95}$ & AP$_{50}$ & AP$_{S}$ & AP$_{M}$ & AP$_{L}$ \\
		\noalign{\smallskip}\hline\noalign{\smallskip}
		YOLOv6-3.0-N & 4.63 & 11.3 & 61.7 & 90.5 & 32.6 & 50.8 & 47.5 \\
		YOLOv7-T & 6.01 & 13.0 & 61.8 & 91.8 & 31.7 & 51.2 & 47.4 \\
		YOLOv8-N & 3.01 & 8.1 & 62.9 & 91.4 & 41.0 & 51.7 & 48.0 \\
		RCS-YOLO & 45.71 & 94.5 & 59.8 & 89.4 & 42.3 & 46.2 & 45.7 \\
		Gold-YOLO-N & 5.60 & 12.1 & 57.0 & 89.7 & 36.6 & 45.9 & 45.5 \\
		YOLOv9-T & 2.62 & 10.7 & 63.0 & 92.2 & 34.8 & 51.8 & 48.1 \\
        YOLOv10-N & 2.70 & 8.2 & 63.0 & 90.8 & 39.7 & 52.5 & 48.5 \\
		\noalign{\smallskip}\hline\noalign{\smallskip}
		DINO-4scale-R50 & 46.60 & 236.1 & 56.8 & 83.4 & 45.1 &43.4 &44.2 \\
		DINO-5scale-R50 & 46.97 & 694.5 & 57.2 & 81.7 & 46.6 &44.4 &44.6 \\
		RT-DETR-R50 & 41.96 & 129.6 & 62.4 & 89.5 &22.3 & 50.2 & 47.8 \\
		RT-DETR-HGNetv2-L & 31.99 &103.4 &61.5 &89.5 &28.7 & 48.7 & 47.0 \\
		\noalign{\smallskip}\hline\noalign{\smallskip}
		LSM-YOLO (ours) & 2.87 & 12.4 & \bfseries{65.1} & \bfseries{92.7} & 41.1 & \bfseries{53.4} & \bfseries{49.7} \\
		\noalign{\smallskip}\hline\hline
	\end{tabular}
\end{table}
\indent Tab.\ref{Tab:3} presents the comparison of our LSM-YOLO with other detectors on the Br35H brain tumor dataset. 
\begin{table}
	\centering
	\caption{Comparison with the state-of-the-art detectors on Br35H dataset.}\label{Tab:3}
	\begin{tabular}{l|c|c|c|c|c|c|c}
		\hline\hline\noalign{\smallskip}	
		Model & Params (M) & GFLOPs & AP$_{50:95}$ & AP$_{50}$ & AP$_{S}$ & AP$_{M}$ & AP$_{L}$ \\
		\noalign{\smallskip}\hline\noalign{\smallskip}
		YOLOv6-3.0-N & 4.63 & 11.3 & 70.7 & 93.2 & 27.6 & 71.4 & 74.5 \\
		YOLOv7-T & 6.01 & 13.0 & 69.7 & 94.5 & 30.6 & 68.9 & 75.8 \\
		YOLOv8-N & 3.01 & 8.1 & 70.0 & 93.6 & 20.7 & 70.6 & 73.5 \\
		RCS-YOLO & 45.70 & 94.5 & 72.9 & 94.6 & 19.8 & 71.5 & 77.3 \\
		Gold-YOLO-N & 5.60 & 12.1 & 70.3 & 93.8 & 24.7 & 71.3 & 73.7 \\
		YOLOv9-T & 2.62 & 10.7 & 71.2 & 93.0 & 18.4 & 70.6 & 76.6 \\
        YOLOv10-N & 2.70 & 8.2 & 69.3 & 91.6 & 20.5 & 68.7 & 72.3 \\
		\noalign{\smallskip}\hline\noalign{\smallskip}
		DINO-4scale-R50 & 46.60 & 203.0 & 68.5 & 91.2 & 21.1 &68.3 &74.8 \\
		DINO-5scale-R50 & 46.97 & 593.8 & 69.7 & 91.9 & 18.2 &70.8 &74.7 \\
		RT-DETR-R50 & 41.96 & 129.5 & 72.1 & 94.1 &27.4 & 70.9 & 75.7 \\
		RT-DETR-HGNetv2-L & 31.99 &103.4 &71.8 &93.0 &29.3 & 69.9 & 76.0 \\
		\noalign{\smallskip}\hline\noalign{\smallskip}
		LSM-YOLO (ours) & 2.87 & 12.4 & \bfseries{73.0} & \bfseries{95.6} & \bfseries{34.5} & 69.9 & 76.7 \\
		\noalign{\smallskip}\hline\hline
	\end{tabular}
\end{table}
Our LSM-YOLO achieves 73.0\% AP$_{50:95}$ and 95.6\% AP$_{50}$, outperforming the previous state-of-the-art model RCS-YOLO while significantly reducing parameter count and computational cost. This establishes LSM-YOLO as the new state-of-the-art model on this dataset.\\
\indent From the AP metrics in Tab.\ref{Tab:3}, LSM-YOLO has a significant advantage in detecting small tumors. This further demonstrates that for the ROI area, our model has effectively learned the contrast features between the target area and its surroundings, resulting in a higher recall rate. It proves that LSM-YOLO can leverage its performance advantages to better assist clinicians in diagnosis.\\
\indent Based on the experimental results from the three datasets, our proposed LSM-YOLO demonstrates its adaptability to detection tasks across multiple organ ROI targets. Compared to the current popular YOLO-series detectors and DETR-series detectors, LSM-YOLO achieves state-of-the-art performance without imposing excessive computational costs. This has evident practical significance for clinical applications.

\subsection{Visualization}
To better demonstrate the detection performance, we use visualization of detection results and Class Activation Map (CAM) to intuitively display the ROI detection effect. In the visualization figures, the first row of sub-figures visualizes the detection results, while the second row represents class activation maps.
\begin{figure}
    \centering
    \subfigure[LSM-YOLO]{
        \includegraphics[width=2.2cm,height=1.5cm]{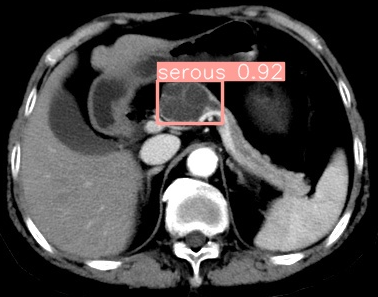} \label{Fig.4(a)}
	}
	\hspace{0.5mm}
    \vspace{-1.0mm}
	\subfigure[RT-DETR]{
		\includegraphics[width=2.2cm,height=1.5cm]{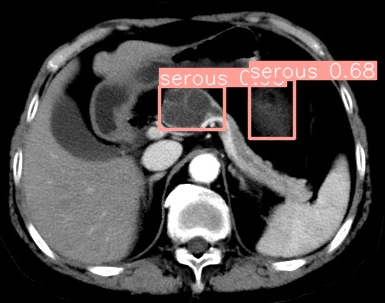} \label{Fig.4(b)}
	}
    \hspace{0.5mm}
    \vspace{-1.0mm}
	\subfigure[YOLOv9]{
		\includegraphics[width=2.2cm,height=1.5cm]{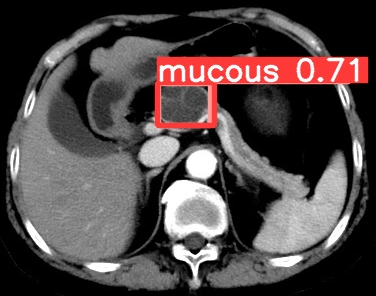} \label{Fig.4(c)}
	}
    \hspace{0.5mm}
    \vspace{-1.0mm}
	\subfigure[YOLOv10]{
		\includegraphics[width=2.2cm,height=1.5cm]{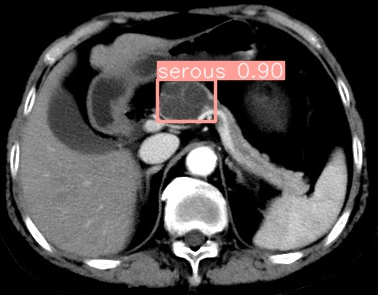} \label{Fig.4(d)}
	}
    
	\subfigure[LSM-YOLO]{
		\includegraphics[width=2.2cm,height=1.5cm]{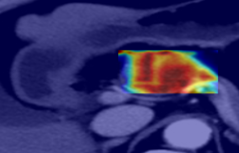} \label{Fig.4(e)}
	}
    \hspace{0.5mm}
    \vspace{-1.0mm}
	\subfigure[RT-DETR]{
		\includegraphics[width=2.2cm,height=1.5cm]{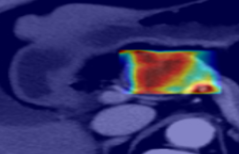} \label{Fig.4(f)}
	}
    \hspace{0.5mm}
    \vspace{-1.0mm}
	\subfigure[YOLOv7]{
		\includegraphics[width=2.2cm,height=1.5cm]{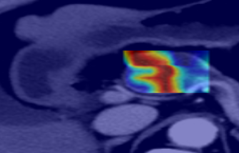} \label{Fig.4(g)}
	}
    \hspace{0.5mm}
    \vspace{-1.0mm}
	\subfigure[YOLOv8]{
		\includegraphics[width=2.2cm,height=1.5cm]{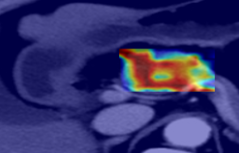} \label{Fig.4(h)}
	}
	\caption{Visual comparisons for tumor ROI detection on the pancreatic tumors dataset.}
	\label{Fig.4}
\end{figure}

\begin{figure}[t]
	\centering
    \subfigure[LSM-YOLO]{
		\includegraphics[width=2.2cm,height=1.5cm]{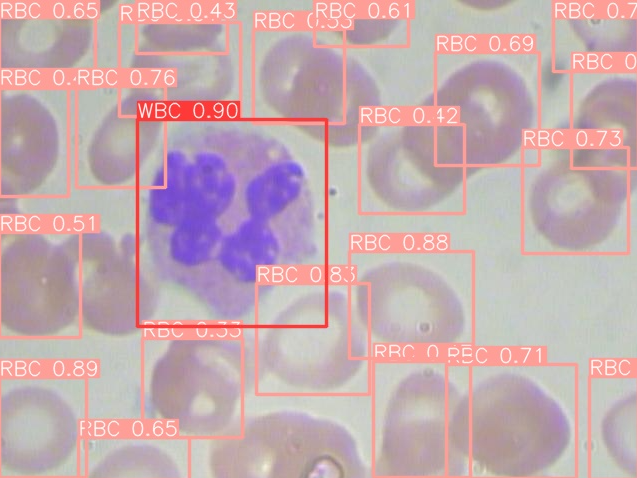} \label{Fig.5(a)}
	}
	\hspace{0.5mm}
    \vspace{-1.0mm}
	\subfigure[RT-DETR]{
		\includegraphics[width=2.2cm,height=1.5cm]{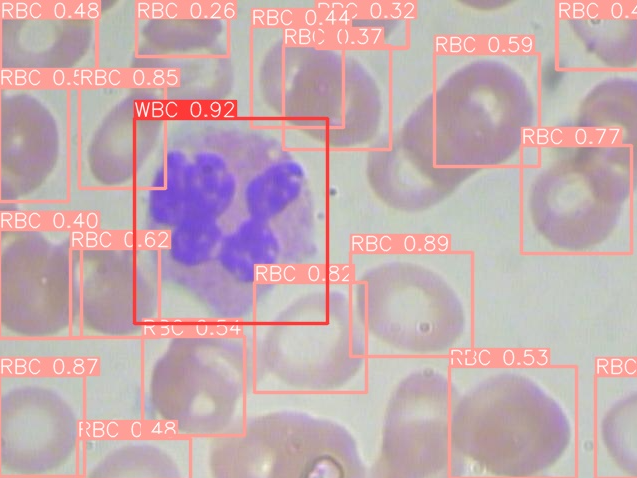} \label{Fig.5(b)}
	}
    \hspace{0.5mm}
    \vspace{-1.0mm}
	\subfigure[YOLOv9]{
		\includegraphics[width=2.2cm,height=1.5cm]{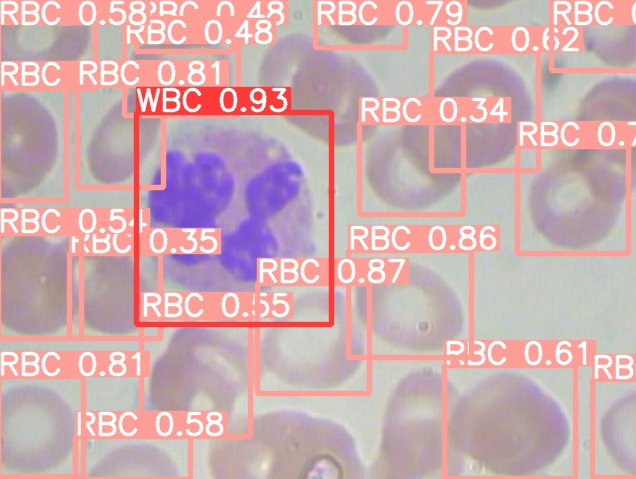} \label{Fig.5(c)}
	}
    \hspace{0.5mm}
    \vspace{-1.0mm}
	\subfigure[YOLOv10]{
		\includegraphics[width=2.2cm,height=1.5cm]{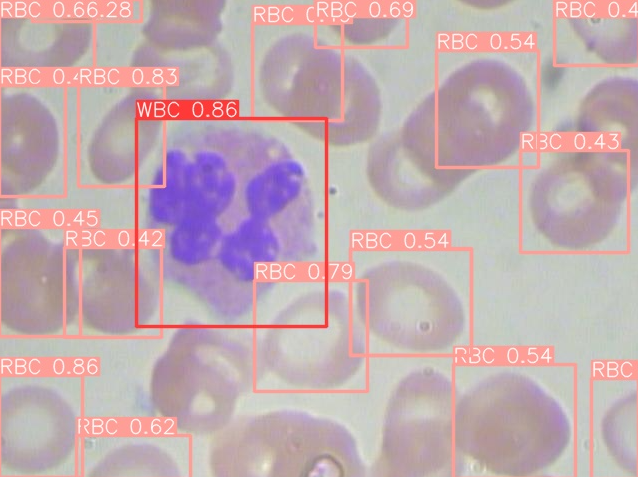} \label{Fig.5(d)}
	}
    
    \subfigure[LSM-YOLO]{
		\includegraphics[width=2.2cm,height=1.5cm]{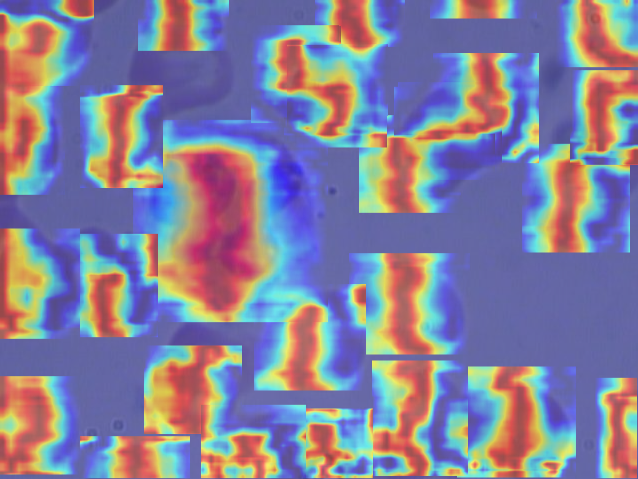} \label{Fig.5(e)}
	}
    \hspace{0.5mm}
    \vspace{-1.0mm}
	\subfigure[RT-DETR]{
		\includegraphics[width=2.2cm,height=1.5cm]{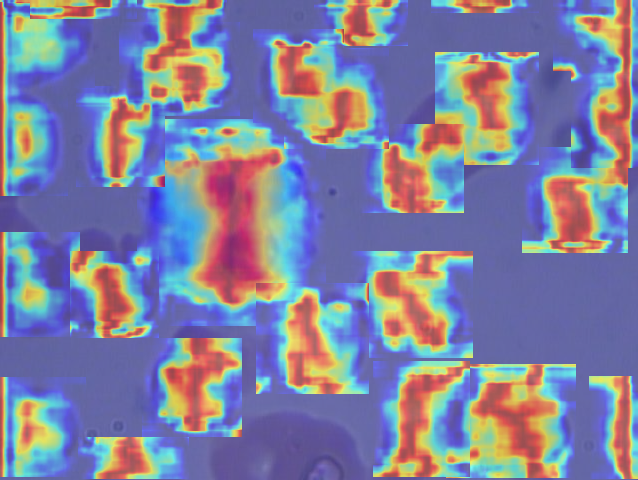} \label{Fig.5(f)}
	}
    \hspace{0.5mm}
    \vspace{-1.0mm}
	\subfigure[YOLOv7]{
		\includegraphics[width=2.2cm,height=1.5cm]{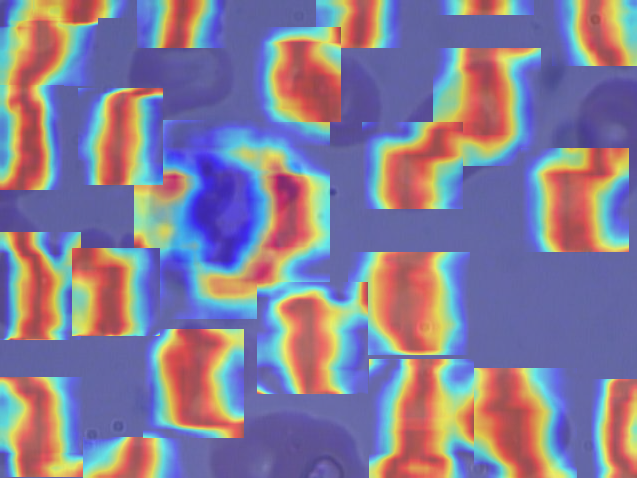} \label{Fig.5(g)}
	}
    \hspace{0.5mm}
    \vspace{-1.0mm}
	\subfigure[YOLOv8]{
		\includegraphics[width=2.2cm,height=1.5cm]{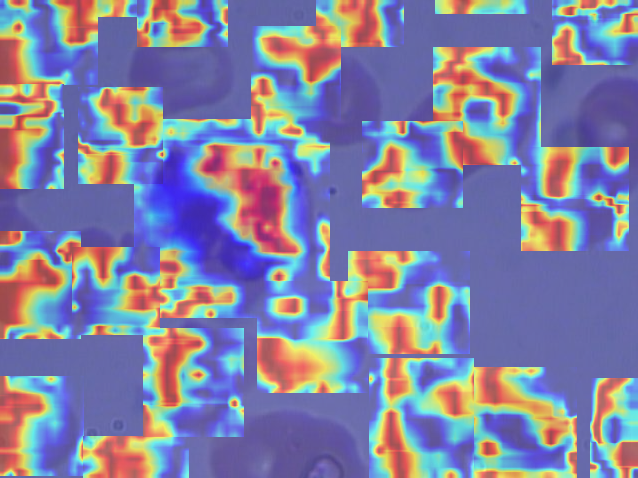} \label{Fig.5(h)}
	}

    \subfigure[LSM-YOLO]{
		\includegraphics[width=2.2cm,height=1.5cm]{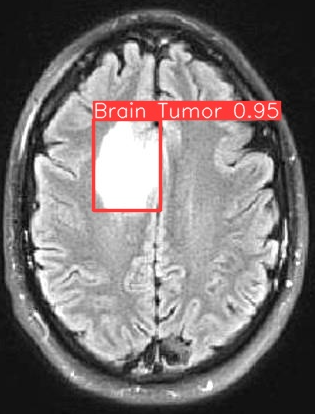} \label{Fig.6(a)}
	}
	\hspace{0.5mm}
    \vspace{-1.0mm}
	\subfigure[RT-DETR]{
		\includegraphics[width=2.2cm,height=1.5cm]{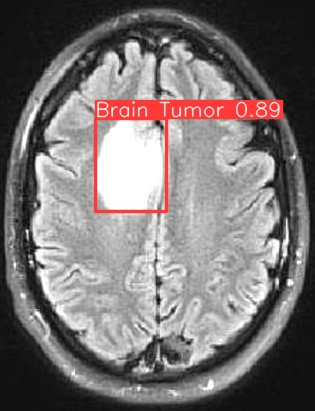} \label{Fig.6(b)}
	}
    \hspace{0.5mm}
    \vspace{-1.0mm}
	\subfigure[YOLOv9]{
		\includegraphics[width=2.2cm,height=1.5cm]{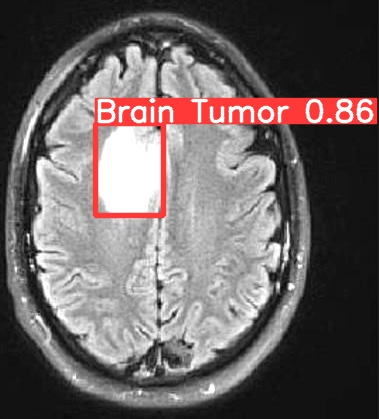} \label{Fig.6(c)}
	}
    \hspace{0.5mm}
    \vspace{-1.0mm}
	\subfigure[YOLOv10]{
		\includegraphics[width=2.2cm,height=1.5cm]{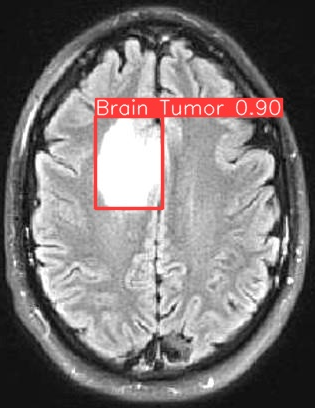} \label{Fig.6(d)}
	}
    
    \subfigure[LSM-YOLO]{
		\includegraphics[width=2.2cm,height=1.5cm]{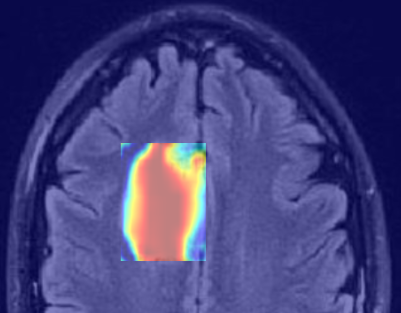} \label{Fig.6(e)}
	}
    \hspace{0.5mm}
    \vspace{-1.0mm}
	\subfigure[RT-DETR]{
		\includegraphics[width=2.2cm,height=1.5cm]{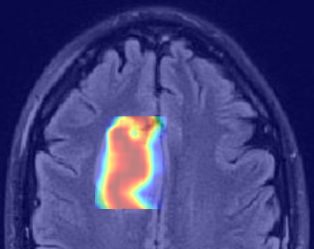} \label{Fig.6(f)}
	}
    \hspace{0.5mm}
    \vspace{-1.0mm}
	\subfigure[YOLOv7]{
		\includegraphics[width=2.2cm,height=1.5cm]{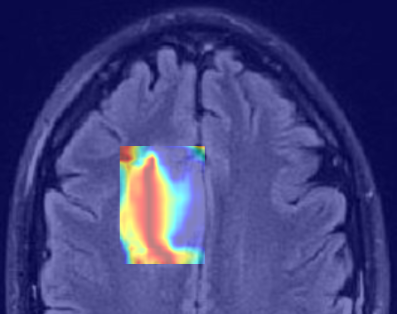} \label{Fig.6(g)}
	}
    \hspace{0.5mm}
    \vspace{-1.0mm}
	\subfigure[YOLOv8]{
		\includegraphics[width=2.2cm,height=1.5cm]{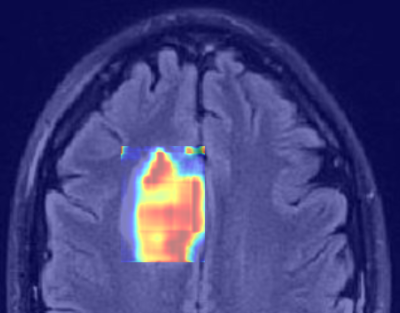} \label{Fig.6(h)}
	}
	\caption{Visual comparisons for cell ROI detection on the BCCD blood cell dataset (upper part) and Br35H brain tumor dataset (lower part).}
	\label{Fig.5}
\end{figure}

\indent From Fig.\ref{Fig.4}'s first row, Fig.\ref{Fig.4(a)} shows LSM-YOLO successfully detecting the tumor location, achieving the highest confidence level of 0.92. Fig.\ref{Fig.4(b)} illustrates a false detection case with RT-DETR, while Fig.\ref{Fig.4(c)} shows that YOLOv9 misclassified the target. In Fig.\ref{Fig.4}'s second row, it can be observed that LSM-YOLO exhibits the best activation effect compared to other detectors. Additionally, it is evident that our LSM-YOLO successfully detects a small portion derived from the right side of the tumor, which other detectors failed to detect.\\
\indent The upper half of Fig.\ref{Fig.5} shows that LSM-YOLO can accurately detect overlapping cells and performs well in dense scenes, a result other detectors cannot achieve. In the class activation maps, LSM-YOLO's activation effect on incomplete cells at the edges is significantly better than that of other detectors.
\indent From the lower half of Fig.\ref{Fig.5}, it can be observed that LSM-YOLO has the highest confidence in correctly classifying brain tumor, and the activation area of the class activation map is the largest. This indicates that LSM-YOLO has the most comprehensive learning of tumor ROI features.

\subsection{Ablation Study}
In order to evaluate the effectiveness of proposed RFABlock, LAE, and MSFM modules, we conduct ablation experiments. As shown in Tab.\ref{Tab:4}, compared to the experiment without these three modules, the addition of RFABlock, LAE, MSFM each improves the detection accuracy respectively. Furthermore, by incorporating the LAE and MSFM modules separately, the experimental results demonstrate that these two modules further enhance the detection performance. Finally, by incorporating all three modules, we obtain the optimal results.

\begin{table}
    \centering
    \caption{Ablation study on proposed RFABlock, LAE, and MSFM modules on the pancreatic tumor dataset.}\label{Tab:4}
    \begin{tabularx}{0.8\textwidth}{Y|Y|Y|Y|Y} 
    \hline\hline\noalign{\smallskip}	
    RFABlock & LAE & MSFM & AP$_{50:95}$ & AP$_{50}$ \\
    \noalign{\smallskip}\hline\noalign{\smallskip}
    &&& 40.3 & 55.0 \\
    \checkmark &&& 42.1 & 58.1 \\
    & \checkmark && 42.7 & 57.6 \\
    && \checkmark & 43.6 & 55.5 \\
    \checkmark && \checkmark & 43.9 & 57.6 \\
    \checkmark & \checkmark && 45.3 & 58.3 \\
    \checkmark & \checkmark & \checkmark & 48.6 & 60.8 \\
    \noalign{\smallskip}\hline\hline
    \end{tabularx}
\end{table}

As shown in Tab.\ref{Tab:5}, we conduct ablation experiments on the internal modules Lightweight Extraction (LE) and Adaptive Extraction (AE) of LAE module, which respectively correspond to the upper and lower parts of Fig.\ref{Fig.2}, and the Dimension Mapping (DM) module. In the experiments without LE, convolution is used for replacement. Incorporating LE and AE into the LAE module enhances the detection rate, with further improvement through DM integration and optimization. This proves the rationality of the internal modules within LAE.

\begin{table}
    \centering 
    \begin{minipage}[t]{0.4\linewidth} 
    \centering 
    \caption{Ablation study on internal modules of LAE on the pancreatic tumor dataset.}
    \label{Tab:5}
    \begin{tabular}{c|c|c|c|c}
        \hline\hline\noalign{\smallskip}
        LE & AE & DM & AP$_{50:95}$ & AP$_{50}$ \\
        \noalign{\smallskip}\hline\noalign{\smallskip}
        &&& 43.7 & 55.0 \\
        \checkmark &&& 45.3 & 56.9 \\
        & \checkmark && 45.1 & 57.5 \\
        & \checkmark & \checkmark & 46.6 & 59.0 \\
        \checkmark & \checkmark & \checkmark & 48.6 & 60.8 \\
        \noalign{\smallskip}\hline\hline
    \end{tabular}
    \end{minipage}
    \hspace{2mm}
    \begin{minipage}[t]{0.4\linewidth} 
    \centering
    \caption{Ablation study on internal modules of MSFM on the pancreatic tumor dataset.}
    \label{Tab:6}
    \begin{tabular}{c|c|c|c}
        \hline\hline\noalign{\smallskip}
        Spatial & Channel & AP$_{50:95}$ & AP$_{50}$ \\
        \noalign{\smallskip}\hline\noalign{\smallskip}
        && 44.8 & 56.5 \\
        \checkmark && 46.7 & 57.2 \\
        & \checkmark & 46.3 & 57.7 \\
        \checkmark & \checkmark & 48.6 & 60.8 \\
        \noalign{\smallskip}\hline\hline
    \end{tabular}
    \end{minipage}
\end{table}
Tab.\ref{Tab:6} displays the ablation experiments of the Spatial and Channel parts within the MSFM, where each part represents the refined fusion of features in different dimensions, which is beneficial for the interaction within the ROI areas and their neighborhoods in medical images. The AP$_{50:95}$ metric improved from 44.8\% to 48.6\%, proving the effectiveness of the two internal components.

\section{Conclusion}
In this paper, we propose the LSM-YOLO network for lightweight and accurate medical ROI detection, which innovatively enhances feature extraction and fusion capabilities,
especially in the detection of small targets in medical images, through the combination
of LAE and MSFM modules. Additionally, the RFABlock is used to expand the receptive field by introducing the attention mechanism into the convolution process. Compared to other object detectors, LSM-YOLO demonstrates superior performance in evaluations on pancreatic tumor datasets, blood cell datasets, and brain tumor datasets, while having a low number of parameters and computational costs, proving its generalization ability and practicality in different medical image detection tasks, suitable for real-time medical image processing. At the same time, we find and verify in experiments that for small-scale medical imaging datasets, large-sized models do not perform as well as small-sized models. In the future, we will continue to explore reliable medical image detectors to assist in clinical diagnosis.

\begin{credits}
\subsubsection{\ackname}
This work is supported in part by National Natural Science Foundation of China (U20A20171, 62373324, 62271448), Natural Science Foundation of Zhejiang Province (LY21F020027) and Key Programs for Science and Technology Development of Zhejiang Province (2022C03113).
\end{credits}

\bibliographystyle{splncs04}
\bibliography{ref}

\end{document}